\documentclass{llncs}
\usepackage{llncsdoc}

\usepackage{graphicx} 
\usepackage{subfigure} 

\usepackage{wrapfig}

\usepackage{psfrag}
\usepackage{amsbsy}
\usepackage[intlimits]{amsmath} 
\usepackage{bbm,amsfonts}

\begin{document}
\markboth{\LaTeXe{} Class for Lecture Notes in Computer
Science}{\LaTeXe{} Class for Lecture Notes in Computer Science}

\pagestyle{empty} 


\title{Feature Selection for Value Function Approximation Using Bayesian Model Selection}
\author{Tobias Jung \and Peter Stone}

\institute{Department of Computer Sciences, University of Texas at Austin, USA\\
{\tt \{tjung,pstone\}@cs.utexas.edu}}

\maketitle

\newcommand{\bx}{\mathbf x}
\newcommand{\bzero}{\boldsymbol 0}
\newcommand{\bv}{\mathbf v}
\newcommand{\bX}{\mathbf X}
\newcommand{\bK}{\mathbf K}
\newcommand{\btheta}{\boldsymbol \theta}

\newcommand{\diag}{\mathop{\mathrm{diag}}}
\newcommand{\trans}{^\text{\sffamily T}}
\newcommand{\norm}[1]{\left\|{#1}\right\|}      
\newcommand{\vect}[1]{\bigl(#1\bigr)\trans}
\newcommand{\br}{\mathbf r}
\newcommand{\bu}{\mathbf u}
\newcommand{\bw}{\mathbf w}
\newcommand{\bH}{\mathbf H}
\newcommand{\boldeta}{\boldsymbol \eta}
\newcommand{\bSigma}{\boldsymbol \Sigma}
\newcommand{\bI}{\mathbf I}
\newcommand{\bQ}{\mathbf Q}
\newcommand{\noise}{\sigma_0^2}
\newcommand{\bxstar}{\bx^*}
\newcommand{\bk}{\mathbf k}
\newcommand{\bmu}{\boldsymbol \mu}

\begin{abstract}   
Feature selection in reinforcement learning (RL), i.e. choosing basis functions
such that useful approximations of the unkown value function can be obtained,
is one of the main challenges in scaling RL to real-world applications. Here we 
consider the Gaussian process based framework GPTD for approximate policy evaluation, and
propose feature selection through marginal likelihood optimization of the associated 
hyperparameters. Our approach has two appealing benefits: (1) given just sample transitions, 
we can solve the policy evaluation problem fully automatically (without looking at the 
learning task, and, in theory, independent of the dimensionality of the state space), and 
(2) model selection allows us to consider more sophisticated kernels, which in turn enable
us to identify relevant subspaces and eliminate irrelevant state variables such that 
we can achieve substantial computational savings and improved prediction performance.
\end{abstract} 

\section{Introduction}
In this paper, we address the problem of approximating the value function under
a stationary policy $\pi$ for a continuous state space $\mathcal X\subset \mathbbm R^D$,
\begin{equation}
V^\pi(\bx)=\mathbbm E_{\bx'|\bx,\pi(\bx)} \ \left\{ R(\bx,\pi(\bx),\bx')+\gamma V^\pi(\bx')\right\}
\label{eq:1.1}
\end{equation}
using a linear approximation of the form 
$
\tilde V(\cdot\,;\bw)=\sum_{i=1}^m w_i \phi_i(\bx)
$
to represent $V^\pi$. Here $\bx$ denotes the state, $R$ the scalar reward and $\gamma$ the discount
factor. Given a trajectory of states $\bx_1,\ldots,\bx_n$ and rewards $r_1,\ldots,r_{n-1}$
sampled under $\pi$, the goal is to determine weights $w_i$ (and basis functions $\phi_i$)
such that $\tilde V$ is a good approximation of $V^\pi$. This is the fundamental problem arising in 
the policy iteration framework of infinite-horizon dynamic programming and reinforcement 
learning (RL), e.g. see \cite{sutton98introduction,bert07dpvolii}.
Unfortunately, this problem is also a very difficult problem that, at present,
has no completely satisfying solution. In particular, deciding which features 
(basis functions $\phi_i$) to use is rather challenging, and in general,
needs to be done manually: thus it is tedious, prone to errors, and most important of all,
requires considerable insight into the domain. Hence, it would be far more desirable if
a learning system could automatically choose its own representation. In particular, considering
efficiency, we want to adapt to the actual difficulties faced, without wasting resources:
often, there are many factors that can make a particular problem easier than it initially appears to be,
for example, when only a few of the inputs are relevant, or when the input data lies on
a low-dimensional submanifold of the input space.

Recent work in applying nonparametric function approximation to RL, such as Gaussian processes (GP) 
\cite{engel2005rlgptd,kuss04gprl,joe2008,deis09gpdp}, or equivalently, regularization networks 
\cite{mein_keepaway2007}, 
is a very promising step in this direction. Instead of having to 
explicitly specify individual basis functions, we only have to specify a 
more general kernel that just depends on a very small 
number of hyperparameters.  
The key contribution of this paper is to demonstrate that 
feature selection in RL from sample transitions can be automated, 
using any of several possible model selection methods for these hyperparameters, such as marginal likelihood optimization 
in a Bayesian setting, or leave-one-out (LOO) error minimization in a frequentist setting. Here, 
we will focus on the Bayesian setting, and adapt marginal likelihood optimization for the GP-based 
approximate policy evaluation method GPTD, introduced without model selection in \cite{engel2005rlgptd}. 
Overall, this will have the following benefits:  
First, only by automatic model selection (as opposed to a grid-based search or manual tweaking of kernel parameters) will we be able to use more sophisticated kernels, which will allow us to uncover the "hidden" properties
of given problem. For example, by choosing an RBF kernel with independent lengthscales for the individual dimensions of the state space, model selection will automatically drive those components to zero that correspond to state variables irrelevant (or redundant) to the task. This will allow us to concentrate our computational efforts
on the parts of the input space that really matter and will improve 
computational efficiency. Second, because it is generally easier to learn in "smaller"
spaces, it may also benefit generalization and thus help us to reduce sample complexity.

Despite its many promises, previous work with GPs in RL rarely explores the benefits 
of model selection: in \cite{joe2008}, a variant of stochastic search was 
used to determine hyperparameters of the covariance for GPTD using as score function the 
online performance of an agent. In \cite{kuss04gprl}, standard GPs with marginal likelihood based model 
selection were employed; however, since their approach was based on fitted value iteration, the task 
of value function approximation was reduced to ordinary regression. 
The remaining paper is structured 
as follows: Section~2-3 contain background information and summarize the GPTD framework. 
As one of the benefits of model selection is the reduction of  computational complexity, 
Section~4 describes how GPTD can be solved for large-scale problems using SR-approximation.
Section~5 introduces model selection for GPTD and derives in detail the associated gradient computation.
Finally, Section~6 illustrates our approach by providing experimental results.   

\section{Related work}
\label{sec:2}
The overall goal of learning representations and feature selection for linearly parameterized 
$\tilde V$ is not new within the context of RL. Roughly, past methods can be categorized
along two dimensions: how the basis functions are represented (e.g. either by parameterized and 
predefined basis functions such as RBF, or by nonparameterized basis functions directly derived 
from the data) and what quantity/target function is 
considered to guide their construction process (e.g. either supervised methods that consider the 
Bellman error and depend on the particular reward/goal, or unsupervised graph-based methods 
that consider connectivity properties of the state space).  
Conceptually closely related to our work is the approach described in \cite{menache05aor},
 which adapts the
hyperparameters of RBF-basis functions (both their location and lengthscales) using 
either gradient descent or the cross-entropy method on the Bellman error. However, because
basis functions are adapted individually (and their number is chosen in advance), 
the method is prone to overfitting: e.g. by placing basis functions 
with very small width near discontinuities. The problem is compounded when only few data
points are available. In contrast, using a Bayesian approach, we can automatically trade-off
model fit and model complexity with the number of data points, choosing always the best 
complexity: e.g. for small data sets we will prefer larger lengthscales (less complex),
for larger data sets we can afford smaller lengthscales (more complex).

Other alternative approaches do not rely on predefined basis functions:
The method in \cite{keller06icml} is an incremental approach that uses dimensionality reduction 
and state aggregation to create new basis functions such that for every step the 
remaining Bellman error for a trajectory of states is successively reduced. A related 
approach is given in \cite{parr07icml} which incrementally constructs an orthogonal basis for the 
Bellman error. A graph-based unsupervised approach is presented in \cite{mahadevan07}, which derives basis 
functions from the eigenvectors of the graph Laplacian induced  from the underlying MDP.

\section{Background: GPs for policy evaluation}
\label{sec:3}
In this section we briefly summarize how GPs \cite{raswil06gp} can 
be used for approximate policy evaluation; here we will follow the GPTD
formulation of \cite{engel2005rlgptd}.

Suppose we have observed 
the sequence
of states $\bx_1,\bx_2,\ldots,\bx_n$ and rewards $r_1,\ldots,r_{n-1}$, where
$\bx_i \sim p(\cdot \, |\, \bx_{i-1},\pi(\bx_{i-1}))$ and 
$r_i=R(\bx_i,\pi(\bx_i),\bx_{i+1})$. In practice, MDPs considered in RL
will often be of an episodic nature with absorbing terminal states. Therefore
we have to transform the problem such that the resulting Markov chain
is still ergodic: this is done by introducing a zero
reward transition from the terminal state of one episode to the start state of the next
episode. In addition to the sequence of states and rewards our training data thus also
includes a sequence $\gamma_1,\ldots,\gamma_{n-1}$, where $\gamma_i=\gamma$ (the discount 
factor in Eq.~\eqref{eq:1.1}) if $\bx_{i+1}$ was a non-terminal state, and $\gamma_i=0$ if $\bx_i$ was 
a terminal state (in which case $\bx_{i+1}$ is the start state of the next episode).

Assume that the function values $V(\bx)$ of the unknown value function $V:\mathcal X \subset
\mathbbm R^D \rightarrow R$ from Eq.~\eqref{eq:1.1} form
a zero-mean Gaussian process with
covariance function $k(\bx,\bx')$ for $\bx,\bx' \in \mathcal X$; in short $V\sim \mathcal{GP}
(\bzero,k(\bx,\bx'))$. In consequence, the function values for the $n$ observed states, 
$\bv:=\vect{V(\bx_1),\ldots,V(\bx_n)}$, will have a Gaussian distribution
\begin{equation}
\bv \,|\, \bX,\btheta \ \sim \ \mathcal N(\bzero,\bK),
\label{eq:2.1}
\end{equation}
where $\bX:=[\bx_1,\ldots,\bx_n]$ and $\bK$ is the $n \times n$ covariance matrix
with entries $[\bK]_{ij}=k(\bx_i,\bx_j)$. Note that the covariance $k(\cdot,\cdot)$
alone fully specifies the GP; here we will assume that it is a simple (positive definite)
function parameterized by a number of scalar parameters collected in vector $\btheta$
(see Section~4).

However, unlike in ordinary regression, in RL we cannot observe samples from the 
target function $V$ directly. Instead, the values can only be observed indirectly: 
from Eq.~\eqref{eq:1.1} we have that the value of one state is recursively defined through the value
of the successor state(s) and the immediate reward. To this end, Engel et al. propose the 
following generative model:\footnote{Note that this model is just a linearly transformed
version of the standard model in GP regression, i.e. $y_i=f(\bx_i)+\varepsilon_i$.}
\begin{equation}
R(\bx_i,\bx_{i+1})=V(\bx_i)-\gamma_i V(\bx_{i+1}) + \eta_i,
\label{eq:2.2}
\end{equation}
where $\eta_i$ is a noise term that may depend on the inputs.\footnote{Formally, in 
GPTD noise is modeled by a second zero-mean GP that is independent from the value GP.
See \cite{engel2005rlgptd} for details.} Plugging in the observed training data, and 
defining $\br:=\vect{r_1,\ldots,r_{n-1}}$, we obtain
\begin{equation}
\br=\bH\bv + \boldeta,
\label{eq:2.3}
\end{equation}
where the $(n-1)\times n$ matrix $\bH$ is given by 
\begin{equation}
\bH:=\begin{bmatrix} 1 & -\gamma_1 &        & \\
               & \ddots    & \ddots & \\
               &           &  1     & -\gamma_{n-1} \end{bmatrix} 
\label{eq:2.4}
\end{equation}
and noise $\boldeta:=\vect{\eta_1,\ldots,\eta_{n-1}}$ has distribution
$\boldeta \sim \mathcal N (\bzero,\bSigma)$. One first choice for the noise 
covariance $\bSigma$ would be $\bSigma=\noise \bI$, where $\noise$ is an unknown 
hyperparameter (see Section~4). However, this model does not capture stochastic
state transitions and hence would only be applicable for deterministic MDPs. If 
the environment is stochastic, the noise model $\bSigma=\noise \bH\bH\trans$ is
more appropriate, see \cite{engel2005rlgptd} for more detailed explanations. For
the remainder we will solely consider the latter choice, i.e. 
$\bSigma=\noise \bH\bH\trans$.

Let $\mathcal D:=\{\bX,\gamma_1,\ldots,\gamma_{n-1}\}$ be an abbreviation
for the training inputs. Using Eq.~\eqref{eq:2.3}, 
it can be shown that the joint distribution
of the observed rewards $\br$ given inputs $\mathcal D$ is again a Gaussian,
\begin{equation}
\br \,|\,\mathcal D,\btheta \ \sim \mathcal N(\bzero,\bQ), 
\label{eq:2.5}
\end{equation}
where the $(n-1)\times(n-1)$ covariance matrix $\bQ$ is given by
\begin{equation}
\bQ=\bigl(\bH\bK\bH\trans + \noise \bH\bH\trans \bigr).
\label{eq:2.6}
\end{equation}

To predict the function value $V(\bxstar)$ at a new state $\bxstar$, we
consider the joint distribution of $\br$ and $V(\bxstar)$
\begin{equation*}
\begin{bmatrix} \br \\ V(\bxstar)
\end{bmatrix}
\,|\,\mathcal D,\bxstar,\btheta \ \sim \
\mathcal N \left(
\begin{bmatrix} \bzero \\ 0 \end{bmatrix},
\begin{bmatrix} \bQ & \bH\bk(\bxstar) \\ 
       [\bH\bk(\bxstar)]\trans & k^*
\end{bmatrix}
\right)
\end{equation*}
where $n \times 1$ vector $\bk(\bxstar)$ is given by 
$\bk(\bxstar):=\vect{k(\bxstar,\bx_1),\ldots,k(\bxstar,\bx_n)}$ and 
scalar $k^*$ by $k^*:=k(\bxstar,\bxstar)$. Conditioning on $\br$, we then 
obtain
\begin{equation}
V(\bxstar)\,|\,\mathcal D,\br,\bxstar,\btheta \ \sim \ 
\mathcal N(\bmu(\bxstar),\sigma^2(\bxstar))
\label{eq:2.8}
\end{equation}
where
\begin{align}
\label{eq:2.9}
\bmu(\bxstar) & := \bk(\bxstar)\trans\bH\trans\bQ^{-1}\br \\
\label{eq:2.10}
\sigma^2(\bxstar) & := k^* -\bk(\bxstar)\trans\bH\trans\bQ^{-1}\bH\bk(\bxstar).
\end{align}
Thus, for any given single state $\bxstar$, GPTD produces the distribution
$p(V(\bxstar)|\mathcal D,\br,\bxstar,\btheta)$ in Eq.~\eqref{eq:2.8} over
function values.

\section{Computational considerations}
\label{sec:Computationalconsiderations}
Regarding its implementation, GPTD for policy evaluation shares the same weakness
that GPs have in traditional machine learning tasks: solving Eq.~\eqref{eq:2.8} requires
the inversion\footnote{For numerical reasons we implement this step using the Cholesky 
decomposition, which has the same computational complexity.} of a dense
$(n-1) \times (n-1)$ matrix, which when done exactly would require $\mathcal O(n^3)$
operations and is hence infeasible for anything but small-scale 
problems (say, anything with $n<5000$).

\newcommand{\bxtilde}{\mathbf{\tilde x}}
\newcommand{\bKtilde}{\mathbf{\tilde K}}
\newcommand{\bW}{\mathbf W}
\newcommand{\bG}{\mathbf G}
\newcommand{\bktilde}{\mathbf{\tilde k}}

\subsection{Subset of regressors}
In the subset of regressors (SR) approach initially proposed for regularization networks
\cite{poggiogirosi90rn,luowahba97has}, one chooses a subset
$\{\bxtilde\}_{i=1}^m$ of the data, with $m\ll n$, and approximates the covariance
for arbitrary $\bx,\bx'$ by taking
\begin{equation}
\tilde k(\bx,\bx')=\bk_m(\bx)\trans\bK_{mm}^{-1}\bk_m(\bx').
\label{eq:x.1}
\end{equation}
Here $\bk_m(\cdot)$ denotes 
$\bk_m(\cdot):=\vect{k(\bxtilde_1,\cdot),\ldots,k(\bxtilde_m,\cdot)}$, and 
$\bK_{mm}$ is the submatrix $[\bK_{mm}]_{ij}=k(\bxtilde_i,\bxtilde_j)$
of 
$\bK$. The approximation in Eq.~\eqref{eq:x.1}
can be motivated for example from the Nystr\"om approximation \cite{williams01nystroem}. 
Let $\bK_{nm}$ denote the 
submatrix $[\bK_{nm}]_{ij}=k(\bx_i,\bxtilde_j)$
corresponding to the $m$ columns of the data points in the subset. We then have 
the rank-reduced approximation $\bK\approx\bKtilde=\bK_{nm}\bK_{mm}^{-1}\bK_{nm}\trans$
and $\bk(\bx)\approx\bktilde(\bx)=\bK_{nm}\bK_{mm}^{-1}\bk_m(\bx)$. Plugging these into 
Eq.~\eqref{eq:2.8}, we obtain for the mean
\begin{align}
\bmu(\bxstar) &\approx \bktilde(\bxstar)\trans\bH\trans
\bigl(\bH\bKtilde\bH\trans+\noise \bH\bH\trans\bigr)^{-1}\br \notag \\
&=\bk_m(\bxstar)\trans\bigl(\bG\trans\bW\bG+\noise \bK_{mm}\bigr)^{-1}\bG\trans\bW\br,
\label{eq:x.2}
\end{align}
where we have defined $\bG:=\bH\bK_{nm}$, $\bW:=(\bH\bH\trans)^{-1}$ and applied the SMW
identity\footnote{
$(\mathbf A+\mathbf B \mathbf D^{-1} \mathbf C)^{-1} \mathbf B \mathbf D^{-1}
=
\mathbf A^{-1}\mathbf B (\mathbf D + \mathbf C\mathbf A^{-1}\mathbf B)^{-1}
$} to show that 
\begin{equation}
\bK_{mm}^{-1}\bG\trans \bigl(\bG\bK_{mm}^{-1}\bG\trans + \noise \bW^{-1}\bigr)^{-1} = 
\bigl(\bG\trans\bW\bG+\noise\bK_{mm}\bigr)^{-1}\bG\trans\bW.
\label{eq:x.3}
\end{equation} 
Similarly, we obtain for the predictive variance
\begin{align}
\sigma(\bxstar) & \approx \tilde k(\bxstar,\bxstar)- 
\bktilde(\bxstar)\trans\bH\trans
\bigl(\bH\bKtilde\bH\trans+\noise \bH\bH\trans\bigr)^{-1}
\bH\bktilde(\bxstar) \notag \\
&=\noise \bk_m(\bxstar)\trans
\bigl(\bG\trans\bW\bG+\noise \bK_{mm}\bigr)^{-1}\bk_m(\bxstar).
\label{eq:x.4}
\end{align}
Doing this means a huge gain in computational savings: solving
the reduced problem in Eq.~\eqref{eq:x.2} costs $\mathcal O(m^2 n)$ for
initialization, requires $\mathcal O(m^2)$ storage and every prediction
costs $\mathcal O(m)$ (or $\mathcal O(m^2)$ if we additionally evaluate the 
variance). This has to be compared with the complexity of the full problem:
$\mathcal O(n^3)$ initialization, $\mathcal O(n^2)$ storage, and $\mathcal O(n)$
prediction. Thus computational complexity now only depends linearly on $n$ (for constant
$m$).

Note that the SR-approximation produces a degenerate GP. As a consequence, the 
predictive variance in Eq.~\eqref{eq:x.4} will underestimate the true variance.
In particular, it will be near zero when $\bx$ is far from the subset $\{\bxtilde\}_{i=1}^m$ (which 
is exactly the opposite of what we want, as the predictive variance should be high 
for novel inputs). The situation can be remedied by considering the projected process
approximation \cite{csato2002neuralcomputation,seeger2003ppa}, which results in the same
expression for the mean in Eq.~\eqref{eq:x.2}, 
but adds the term 
\begin{equation}
k(\bxstar,\bxstar)-\bk_m(\bxstar)\trans\bK_{mm}^{-1}\bk_m(\bxstar)
\label{eq:x.5}
\end{equation}
to the variance in Eq.~\eqref{eq:x.4}

\subsection{Selecting the subset (unsupervised)}
Selecting the best subset 
is a combinatorial problem that cannot be solved effeciently. Instead, we try to find
a compact subset that summarizes the relevant information by incremental forward selection. 
In every step of the procedure, we add that element from the set of remaining unselected 
elements to the active set that performs best with respect to a given specific criterion.
In general, we distinguish between supervised and unsupervised approaches, i.e. those 
that consider the target variable we regress on, and those that do not. Here we focus on 
the incomplete Cholesky decomposition (ICD) as an unsupervised approach 
\cite{fine01efficientsvmtrainin,bach02incompletecholesky,bach05icdsupervised}.

ICD aims at reducing at each step the error incurred from approximating the
covariance matrix: $\norm{\bK-\bKtilde}_F$. Note that the ICD of $\bK$ is the dual
equivalent of performing partial Gram-Schmidt on the Mercer-induced feature representation:
in every step, we add that element to the active set whose distance from the span
of the currently selected elements is largest (in feature space). The procedure is stopped
when the residual of remaining (unselected) elements falls below a given threshold,
or a given maximum number of allowed elements is exceeded. In 
\cite{csato2002neuralcomputation,mein_keepaway2007,engel2005rlgptd} online variants 
thereof are considered (where instead of repeatedly inspecting all remaining elements
only one pass over the dataset is made and every element is examined only once). In general,
the number of elements selected by ICD will depend on the effective rank of $\bK$ 
(and thus its eigenspectrum).

\section{Model selection for GPTD}
The major advantage of using GP-based function approximation
(in contrast to, say, neural networks or tree-based approaches) is that both
'learning' of the weight vector and specification of the architecture/hyperparameters/basis 
functions can be handled in a principled and essentially automated way.

\subsection{Optimizing the marginal likelihood}
To determine hyperparameters for GPTD, we consider the marginal likelihood of the process, i.e.
the probability of generating the rewards we have observed given the sequence of states 
and a particular setting of the hyperparameter $\btheta$. We then maximize this function
(its logarithm) with respect to $\btheta$. From Eq.~\eqref{eq:2.5} we see that for GPTD we
have $p(\br|\mathcal D,\btheta)=\mathcal N(\bzero,\bQ)$. Thus plugging in the definition for
a multivariate Gaussian and taking the logarithm, we obtain
\begin{equation}
\mathcal L(\btheta)
=
-\frac{1}{2} \log\det\bQ 
-\frac{1}{2}\br\trans\bQ^{-1}\br  
-\frac{n}{2} \log 2\pi.
\label{eq:3.1}
\end{equation}
Optimizing this function with respect to $\btheta$ is a nonconvex problem and we have to resort
to iterative gradient-based solvers (such as scaled conjugate gradients, e.g. see 
\cite{nabney02netlab}). To do this we need to be able to evaluate the gradient of $\mathcal L$.
The partial derivatives of $\mathcal L$ with respect to each individual hyperparameter $\theta_i$
can be obtained in closed form as
\begin{equation}
\frac{\partial\mathcal L}{\partial \theta_i} =
-\frac{1}{2} \mathrm{tr}\left(\bQ^{-1}\frac{\partial\bQ}{\partial \theta_i}\right)
+\frac{1}{2} \br\trans\bQ^{-1}\frac{\partial\mathcal \bQ}{\partial \theta_i}\bQ^{-1}\br.
\label{eq:3.2}
\end{equation}
Note that $\mathcal L$ automatically incorporates the trade-off between model fit
(training error) and model complexity and can thus be regarded as an indicator for generalization
capabilities, i.e. how well GPTD will predict the values of states not in its training set. The
first term in Eq.~\eqref{eq:3.1} measures the complexity of the model, and will be large for 'flexible'
and small for 'rigid' models.\footnote{A property that manifests itself in the eigenvalues
of $\bK$ (since the determinant equals the sum of the eigenvalues). In general, flexible models
are achieved by smaller bandwidths in the covariance, meaning that $\bK$'s effective rank will be
large and its eigenvalues will fall off more slowly. On the other hand, more rigid
models are achieved by larger bandwidths, meaning that $\bK$'s effective rank will be low and its eigenvalues will fall off more quickly. Note that the effective rank of $\bK$ is 
also important for the SR-approximation (see Section~3), since the effectiveness of SR depends on
building a low-rank approximation of $\bK$ spending as few resources as possible.} The second term
measures the model fit and can shown to be the value of the error function for a penalized least-squares
that would (in a frequentist setting) correspond to GPTD.

\newcommand{\bOmega}{\boldsymbol \Omega}
\newcommand{\bM}{\mathbf M}
\newcommand{\bm}{\mathbf m}

\subsection{Choosing the covariance}
\label{sec:choosingthecovariance}
\begin{figure}[tb]
\psfrag{h}{\scriptsize $h$}
\psfrag{a1}{\scriptsize $a_1$}
\psfrag{a2}{\scriptsize $a_2$}
\psfrag{s1}{\scriptsize $s_1$}
\psfrag{s2}{\scriptsize $s_2$}
\psfrag{u1}{\scriptsize $\mathbf u_1$}
\psfrag{u2}{\scriptsize $\mathbf u_2$}
\begin{center}
\includegraphics[width=0.99\textwidth]{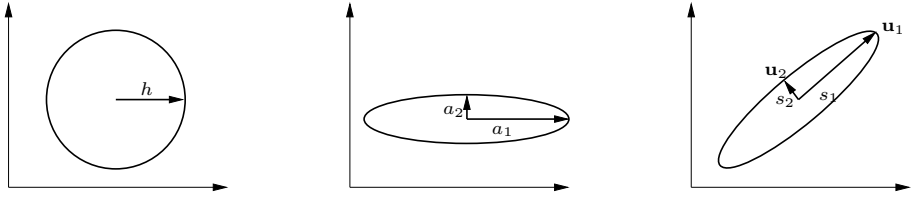}
\end{center}
\caption{Three variants of the stationary squared exponential covariance. The
directions/scaling factors in the third case are derived from the eigendecomposition 
of $\bOmega$, i.e. $\mathbf{USU}\trans=\bM_k\bM_k\trans + \diag(a_1,\ldots,a_D)$.} 
\label{fig:1}
\end{figure}

A common choice for $k(\cdot,\cdot)$ is to consider a (positive definite) function
parameterized by a small number of scalar parameters, such as the stationary isotropic
Gaussian (or squared exponential), which is parameterized by the lengthscale (bandwidth $h$).
In the following we will consider three variants of the form \cite{nabney02netlab,raswil06gp}:
\begin{equation}
k(\bx,\bx')=v_0 \exp\left\{-\frac{1}{2}(\bx-\bx')\trans\bOmega(\bx-\bx')\right\}+b
\label{eq:3.3}
\end{equation}
where hyperparameter $v_0>0$ denotes the vertical lengthscale, $b>0$ the bias, and symmetric
positive semidefinite matrix $\bOmega$ is given by
\begin{itemize}
\item {\bf Variant 1 (isotropic):} 
$ \bOmega=h\bI $ \newline
with hyperparameter $h>0$.
\item {\bf Variant 2 (axis-aligned ARD):}
$ \bOmega=\diag(a_1,\ldots,a_D) $ \newline
with hyperparameters $a_1,\ldots,a_D>0$.
\item {\bf Variant 3(factor analysis):}
$ \bOmega=\bM_k\bM_k\trans + \diag(a_1,\ldots,a_D) $ \newline
where $D \times k$ matrix $\bM_k$ is given by $\bM_k:=[\bm_1,\ldots,\bm_k]$, $k<D$, and 
both the entries of $\bM_k$, i.e. $m_{11},\ldots,m_{1D},\ldots,m_{k1},\ldots,m_{kD}$ 
and $a_1,\ldots,a_D>0$ are adjustable hyperparameters.\footnote{The number of directions
$k$ is also determined from model selection: we systematically try different values of $k$,
find the corresponding remaining hyperparameters via scg-based likelihood optimization
and compare the final scores (likelihood) of the resulting models.}
\end{itemize}
The first variant (see Figure~\ref{fig:1}) assumes that every coordinate of the input (i.e. state-vector)
is equally important for predicting its value. However, in particular for high-dimensional state vectors, this might be too simple: along some dimensions this will produce too 
much resolution where it will be wasted, along other dimensions this will produce too little resolution
where it would otherwise be needed. The second variant is more powerful and includes
a different parameter for every coordinate of the state vector, thus assigning a different scale
to every state variable. This covariance implements automatic relevance determination
(ARD): since the individual scaling factors are automatically adapted from the data via marginal 
likelihood optimization, they inform us about how relevant each state variable is for predicting
the value. A large value of $a_i$ means that the $i$-th state variable is important and even small
variations along this coordinate are relevant. A small value of $a_j$ means that the $j$-th 
state variable is less important and only large variations along this coordinate will impact the
prediction (if at all). A value close to zero means that the corresponding coordinate is
irrelevant and could be left out (i.e. the value function does not rely on that particular
state variable). The benefit of removing irrelevant coordinates is that the complexity of the
model will decrease while the fit of the model stays the same: thus likelihood will increase. The 
third variant first identifies relevant directions in the input space (linear 
combinations of state variables) and performs a rotation of the coordinate system (the number of 
relevant directions is specified in advance by $k$). As in the second variant, different scaling 
factors are then applied along the rotated axes.

\newcommand{\bC}{\mathbf C}
\newcommand{\one}{\mathbbm 1_{n,n}}

\subsection{Example: gradient for ARD}
As an example, we will now show how the gradient $\nabla_{\btheta} \mathcal L$ of Eq.~\eqref{eq:3.1}
is calculated for the ARD covariance. 
Note that since all hyperparameters 
in this model, i.e. $\{v_0,b,\noise,a_1,\ldots,a_D\}$, must be positive, it is more convenient
to consider the hyperparameter vector $\btheta$ in log space: $\btheta=\bigl( \log v_0,
\log b,\log \noise, \log a_1,\ldots,\log a_D \bigr)$. 
We start by establishing some useful identities: for any $n \times n$ matrix $\mathbf A$
we have
\begin{equation*}
[\bH\mathbf A\bH\trans]_{ij}=a_{ij}-\gamma_i a_{i+1,j}-\gamma_j a_{i,j+1} 
+\gamma_i \gamma_j a_{i+1,j+1}.
\end{equation*}
Furthermore, we have
\begin{equation*}
[\bH\bH\trans]_{ij}=\begin{cases} 1+\gamma_i^2 & ,i=j \\
                                  -\gamma_i    & ,i=j-1 \textrm{ or } i=j+1\\
                                  0 &, \textrm{ otherwise}
                     \end{cases}
\end{equation*}
Now write $\bK$ as $\bK=v_0\bC+b\one$, where $[\bC]_{ij}=\exp\left\{-0.5 \sum_{d=1}^D
a_d\bigl(x_d^{(i)} - x_d^{(j)}\bigr)^2\right\}$ and $\one$ is the $n \times n$ matrix
of all ones. Computing the partial derivative of $\bK$, we then obtain
\begin{equation*}
\frac{\partial \bK}{\partial v_0}=\bC , \quad 
\frac{\partial \bK}{\partial b}=\one \\
\end{equation*}
\begin{equation*}
\left[ \frac{\partial \bK}{\partial a_\nu}\right]_{ij}=-\frac{1}{2}v_0c_{ij}
\bigl(x_\nu^{(i)} - x_\nu^{(j)}\bigr)^2, \quad \nu=1\ldots D
\end{equation*} 
Next, we will compute the partial derivatives of $\bQ=(\bH\bK\bH\trans+\noise\bH\bH\trans)$,
giving for $b$:
\begin{equation*}
\begin{split}
&\frac{\partial \bQ}{\partial \log b}=b\frac{\partial \bQ}{\partial b}=
b\bH \left[ \frac{\partial \bK}{\partial b}\right] \bH\trans=
b\bH\one\bH\trans \\
&\Rightarrow \left[\frac{\partial \bQ}{\partial \log b}\right]_{ij}=
b(1-\gamma_i-\gamma_j+\gamma_i\gamma_j).
\end{split}
\end{equation*}
For $v_0$ we have 
\begin{equation*}
\begin{split}
\frac{\partial \bQ}{\partial \log v_0}&=v_0\frac{\partial \bQ}{\partial v_0}=
v_0\bH \left[ \frac{\partial \bK}{\partial v_0}\right] \bH\trans=
v_0\bH\bC\bH\trans \\
\Rightarrow \left[\frac{\partial \bQ}{\partial \log v_0}\right]_{ij}&=
v_0(c_{ij}-\gamma_ic_{i+1,j}- 
\gamma_jc_{i,j+1}+\gamma_i\gamma_jc_{i+1,j+1}).
\end{split}
\end{equation*}
For $\noise$ we have
\begin{equation*}
\begin{split}
&\frac{\partial \bQ}{\partial \log \noise}=\noise\frac{\partial \bQ}{\partial \noise}=
\noise \frac{\partial}{\partial \noise} [\noise \bH\bH\trans]=\noise\bH\bH\trans\\
&\Rightarrow \left[\frac{\partial \bQ}{\partial \log \noise}\right]_{ij}=
\begin{cases} \noise(1+\gamma_i^2)& ,i=j \\
                                  -\noise\gamma_i    & ,i=j-1 \textrm{ or } i=j+1\\
                                  0 &, \textrm{ otherwise}
                     \end{cases}
\end{split}
\end{equation*}
Finally, for each of the $a_\nu$, $\nu=1,\ldots,D$ we get
\begin{equation*}
\begin{split}
\frac{\partial \bQ}{\partial \log a_\nu}&=a_\nu\frac{\partial \bQ}{\partial a_\nu}=
a_\nu \bH \left[\frac{\partial \bK}{\partial a_\nu}\right]\bH\trans \\
\Rightarrow\left[ \frac{\partial \bQ}{\partial \log a_\nu}\right]_{ij}&=-\frac{1}{2}
a_\nu v_0 (c_{ij}d_{ij}^\nu-\gamma_i c_{i+1,j}d_{i+1,j}^\nu \\
&-\gamma_j c_{i,j+1}d_{i,j+1}^\nu + \gamma_i \gamma_j c_{i+1,j+1} d_{i+1,j+1}^\nu)
\end{split}
\end{equation*}
where we have defined $d_{ij}^\nu:=\bigl(x_\nu^{(i)} - x_\nu^{(j)}\bigr)^2$.
Thus, with $\bw:=\bQ^{-1}\br$ we have for Eq.~\eqref{eq:3.2}
\begin{align*}
\textrm{tr}\left(\bQ^{-1}\frac{\partial \bQ}{\partial \theta_\nu}\right)
& =
\sum_{i=1}^{n-1} \sum_{j=1}^{n-1} \left[\bQ^{-1}\right]_{ij}
\left[\frac{\partial \bQ}{\partial \theta_\nu}\right]_{ji} \\
\bw\trans \frac{\partial \bQ}{\partial \theta_\nu}\bw 
& =
\sum_{i=1}^{n-1} \sum_{j=1}^{n-1} [\bw]_i [\bw]_j
\left[\frac{\partial \bQ}{\partial \theta_\nu}\right]_{ij}
\end{align*}
which can be used to calculate the partial derivates 
with computational complexity $\mathcal O(n^2)$ each (except for $\noise$, 
where the matrix of derivatives is tridiagonal).

\section{Experiments}
This section demonstrates that our proposed model selection can be used to solve the 
approximate policy evaluation problem in a completely automated way -- 
without any manual tweaking of hyperparameters.
We will also show some of the additional benefits of model selection, which are  
improved accuracy and reduced complexity: because we automatically
set the hyperparameters we can use more sophisticated covariance functions (see 
Section~\ref{sec:choosingthecovariance}) that depend on a larger number\footnote{Setting
these hyperparameters by hand would require even more trial and error; therefore,
these covariances are seldom employed without model selection.} of hyperparameters, thus
better fit the regularities of a particular dataset, and therefore do not waste unnecessary
resources on irrelevant aspects of the state-vector. 
The latter aspect is particularly
interesting for computational reasons (see Section~4) and becomes important in 
large-scale applications.  

\begin{wrapfigure}{r}{0.5\textwidth}
\begin{center}
\includegraphics[width=0.5\textwidth]{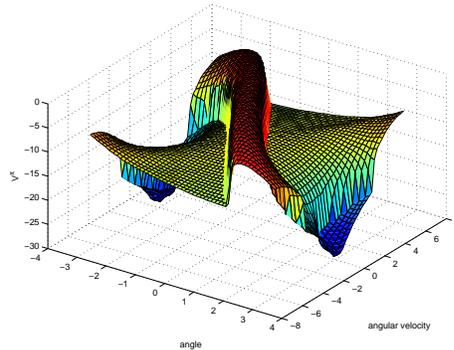}
\end{center}
\caption{Optimal value function for the pendulum domain, computed with fitted value
iteration over a discretized state space ($400 \times 400$ grid).} 
\label{fig:2a}
\end{wrapfigure}

\subsection{Pendulum swing-up task}
\begin{figure}[p]
\begin{center}
\includegraphics[width=0.55\textwidth]{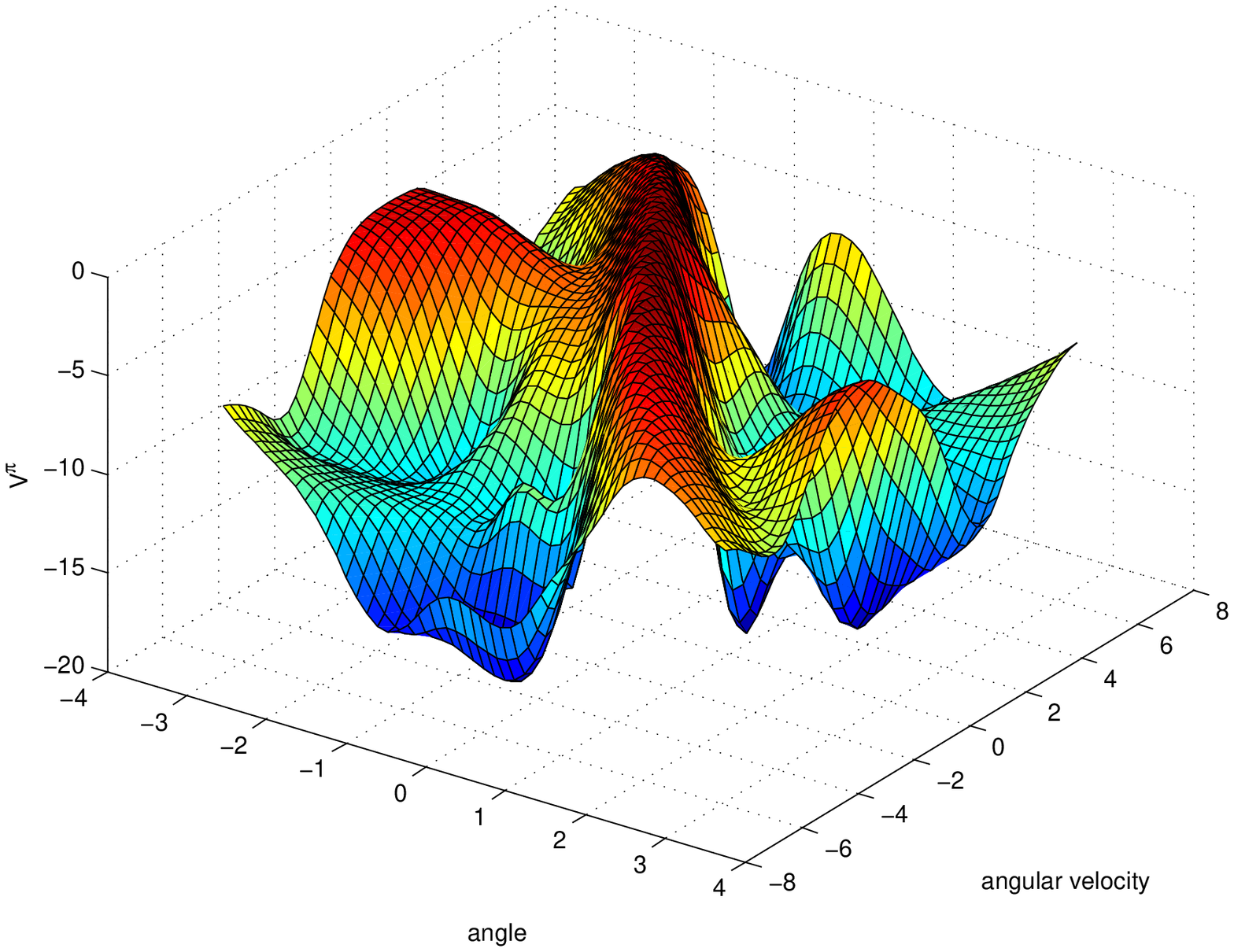}
\includegraphics[width=0.44\textwidth]{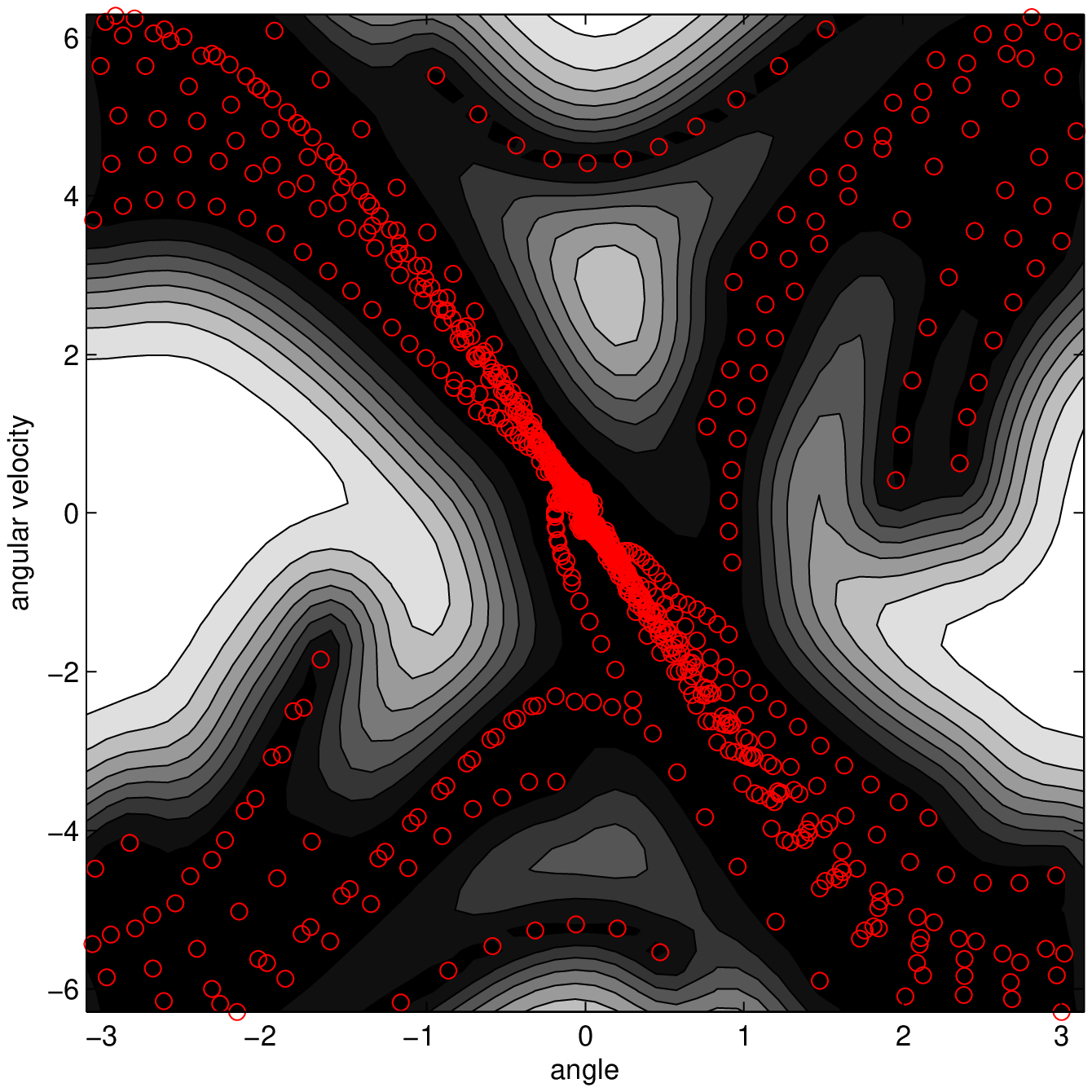}
\includegraphics[width=0.55\textwidth]{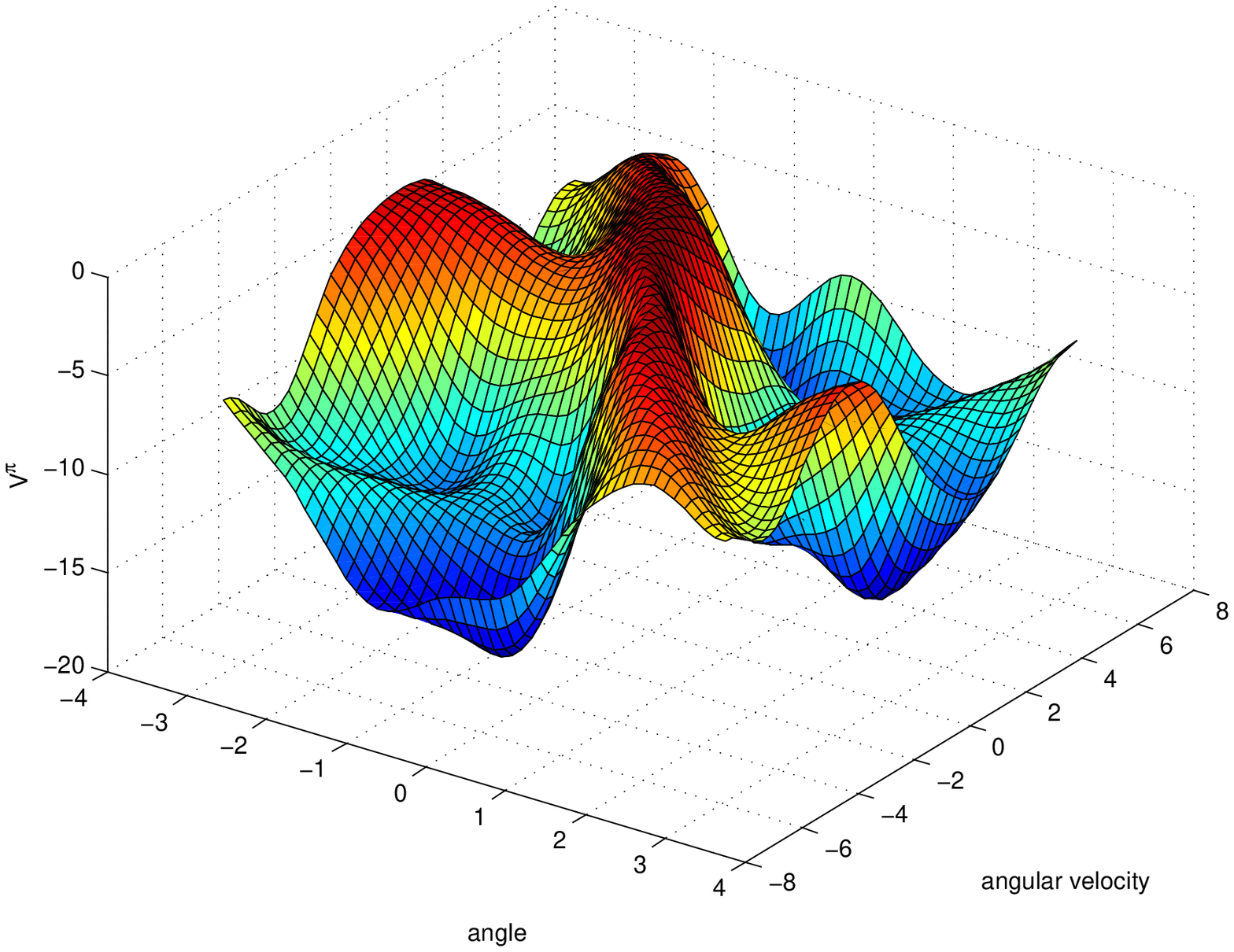}
\includegraphics[width=0.44\textwidth]{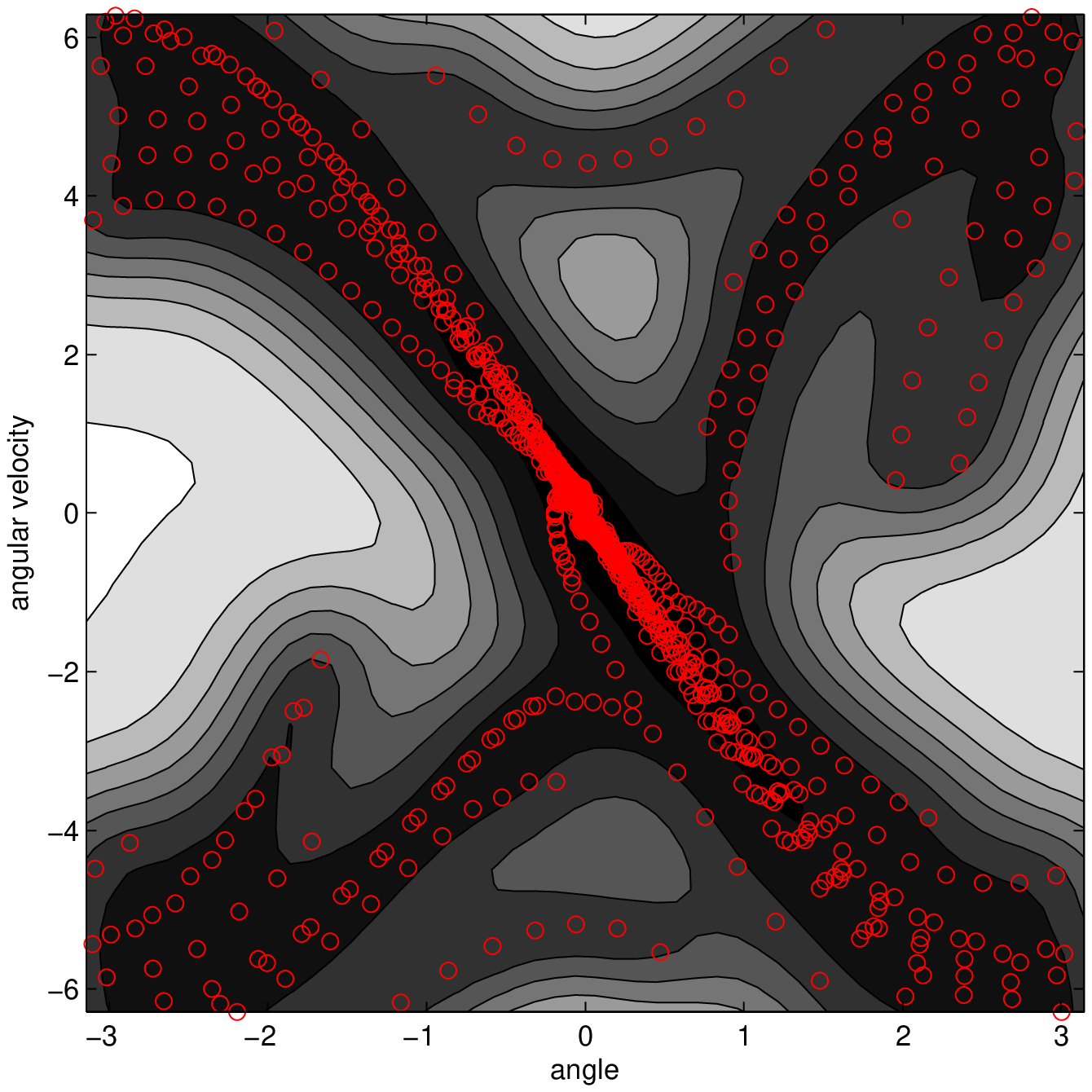}
\includegraphics[width=0.55\textwidth]{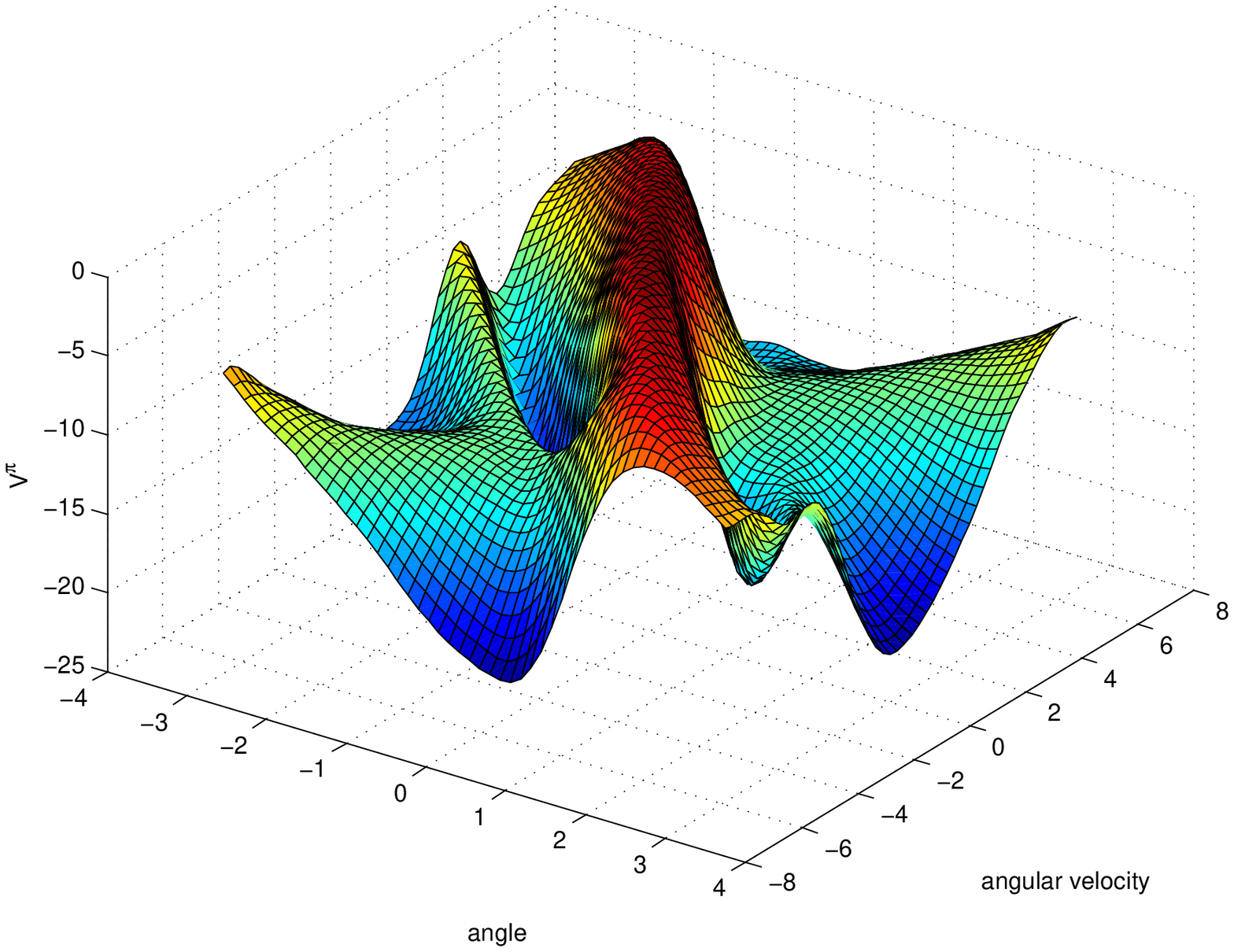}
\includegraphics[width=0.44\textwidth]{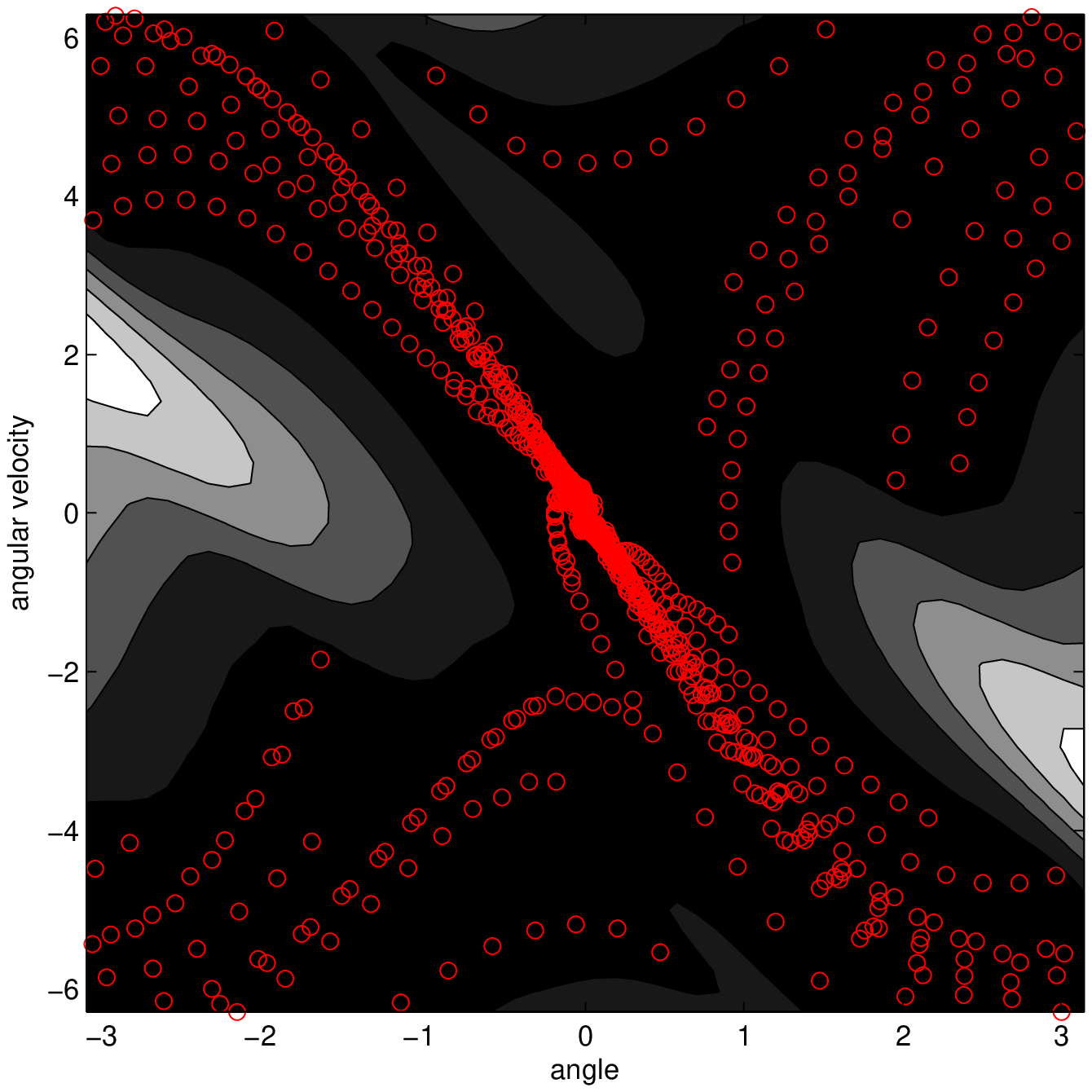}
\end{center}
\caption{From top to bottom: GPTD approximation of the value function from Figure~\ref{fig:2a} 
for the covariances (I),(II),(III), where in each case the hyperparameters were obtained
from marginal likelihood optimization for the GPTD process in Eq.~\eqref{eq:3.1}. 
Right: Associated predictive variance. Black
indicates low variance, white indicates high variance and red circles indicate the
location of the states in the training set (which was the same for all three experiments).} 
\label{fig:2}
\end{figure}

First, we consider the pendulum swing-up task, a common benchmark in RL.
The goal is to swing up an underpowered pendulum and balance it around the 
inverted upright position (here formulated as an episodic task). More details
and the equations of motion can be found in e.g. \cite{deis09gpdp}.  
Since GPTD only solves (approximate) policy evaluation, to test our model selection
approach we chose to generate a sample trajectory under the optimal policy (obtained
from fitted value iteration). We generated a sequence of 1000 state-transitions under 
this policy (which corresponds to about 25 completed episodes) and applied GPTD for 
the three choices of covariance: isotropic (I), axis-aligned ARD (II), and factor 
analyis (III). In each case, the best setting of hyperparameters was found from 
running\footnote{We used the full data set for model selection, to avoid 
the complexities involved with subset-based likelihood approximation, 
e.g. see \cite{spgp2006}. In our implementation, model selection for 
all 1000 data points took
about 15-30 secs on a 1.5GHz PC.} scaled conjugate gradients on 
Eq.~\eqref{eq:3.1}, giving 
\begin{center}
{\scriptsize
\begin{tabular}{ccccccc}
I: $v_0=18.19$&$\sigma_0^2=0.05$ & $b=0.11$ & $h=7.48$ & & \\
II: $v_0=15.95$&$\sigma_0^2=0.05$ & $b=0.10$ & $a_1=3.62$ & $a_2=6.63$ & & \\
III: $v_0=10.82$&$\sigma_0^2=0.08$ & $b=0.10$ & $s_1=13.91$ & $s_2=0.36$ & 
$\bu_1=\begin{bmatrix} 0.58 & 0.81 
\end{bmatrix}$
 & $\bu_2=\begin{bmatrix} -0.81 & 0.58 \end{bmatrix}$ \\
\end{tabular}
}
\end{center}
\noindent
(the last ones given in terms of the eigendecomposition of $\bOmega$). Figure~\ref{fig:2}
shows the results: all three produce an adequate representation of the true value function shown in 
Figure~\ref{fig:2a} in and near
the states visited in the trajectory (MSE in states of the sample 
trajectory: (I) 0.27, (II) 0.24, and (III) 0.26), but differ once they start
predicting values of states not in the training data (MSE for states on a $50 \times 50$ grid: 
(I) 46.36, (II) 48.89, and (III) 12.24). 
Despite having a slightly higher error on the known training data, (III) substantially 
outperforms the other models when it comes to predicting the values of new states. With respect 
to model selection, (III) also has the highest likelihood. Note that 
(III) chooses one dominant direction ($\bu_1=\begin{bmatrix} 0.58 & 0.81 
\end{bmatrix}$) to which it assigns high relevance ($s_1=13.91$); the remainder
($\bu_2=\begin{bmatrix} -0.81 & 0.58 \end{bmatrix}$) has only  
little impact ($s_2=0.36$). Taking a closer look at Figure~\ref{fig:2a}, we see that indeed the value 
function varies more strongly along the diagonal direction lower left to upper right, whereas
it varies only slowly along the opposite diagonal upper left to lower right.
For (II), relevance can only be assigned along the
$\varphi$ and $\dot\varphi$ coordinates (state-variables), which in this case gives 
us no particular benefit; and (I) is not at all able to assign different
importance to different state variables.

\begin{figure}[tb]
\begin{center}
\includegraphics[width=0.32\textwidth, height=4cm]{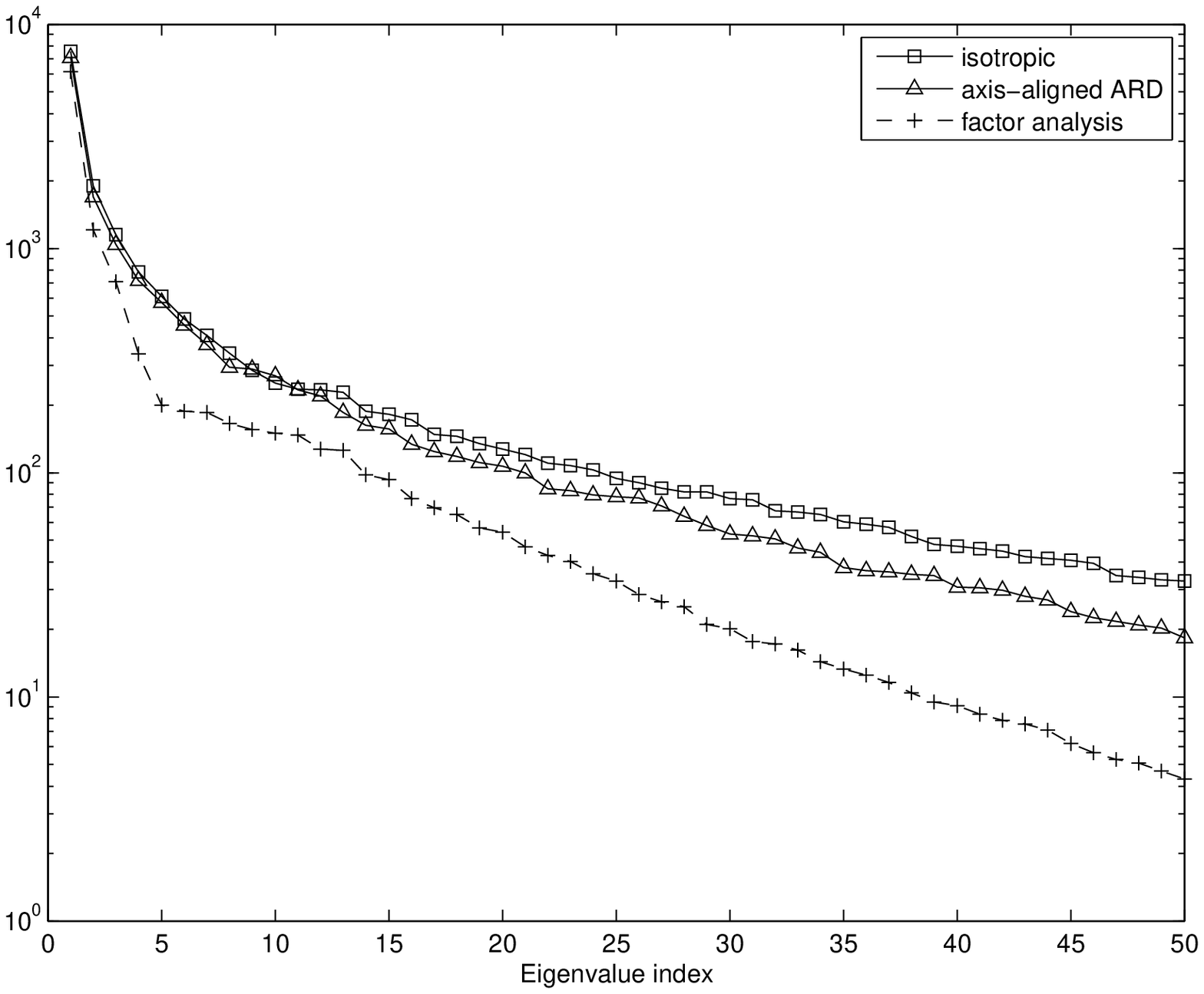}
\includegraphics[width=0.32\textwidth, height=4cm]{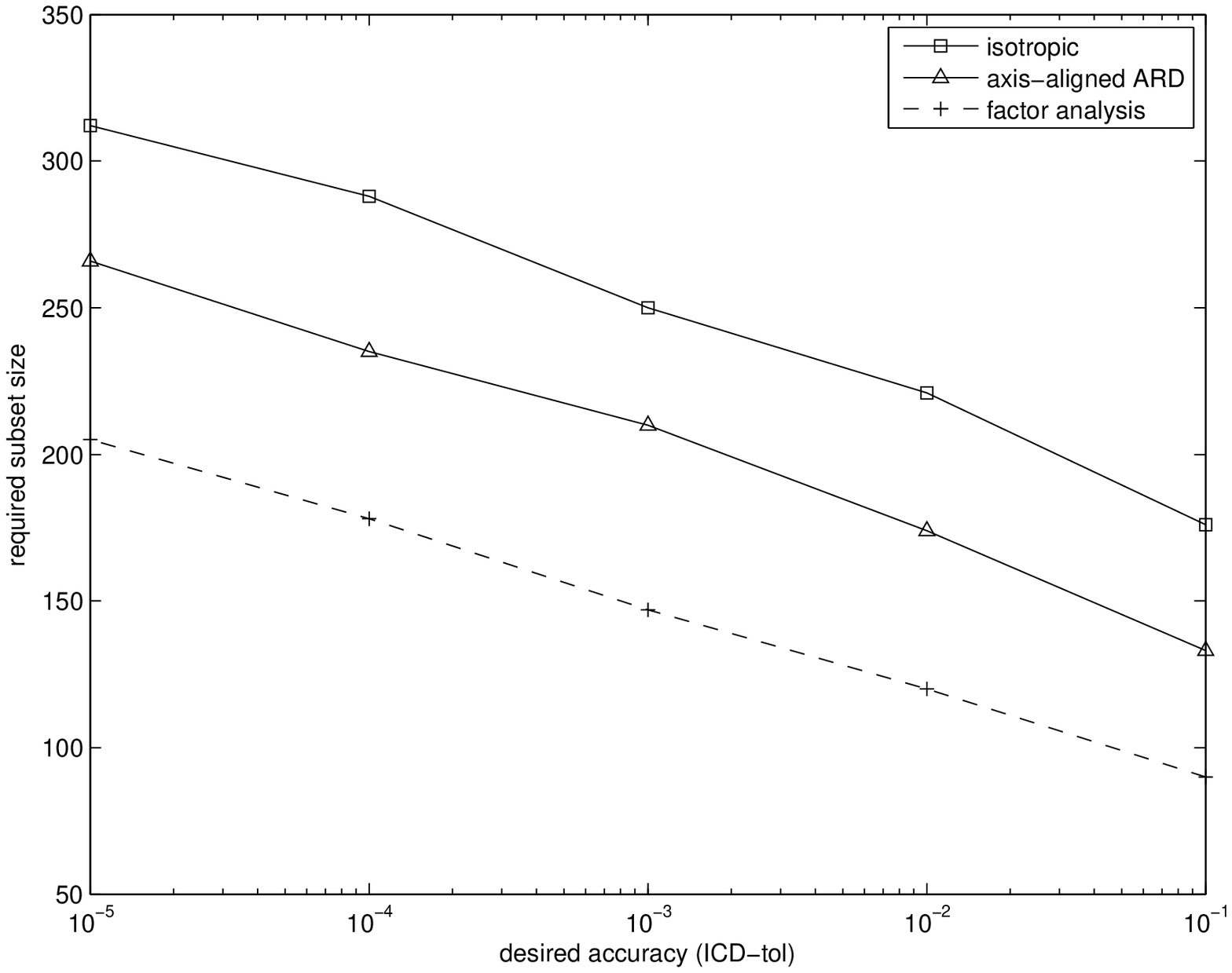}
\includegraphics[width=0.32\textwidth, height=4cm]{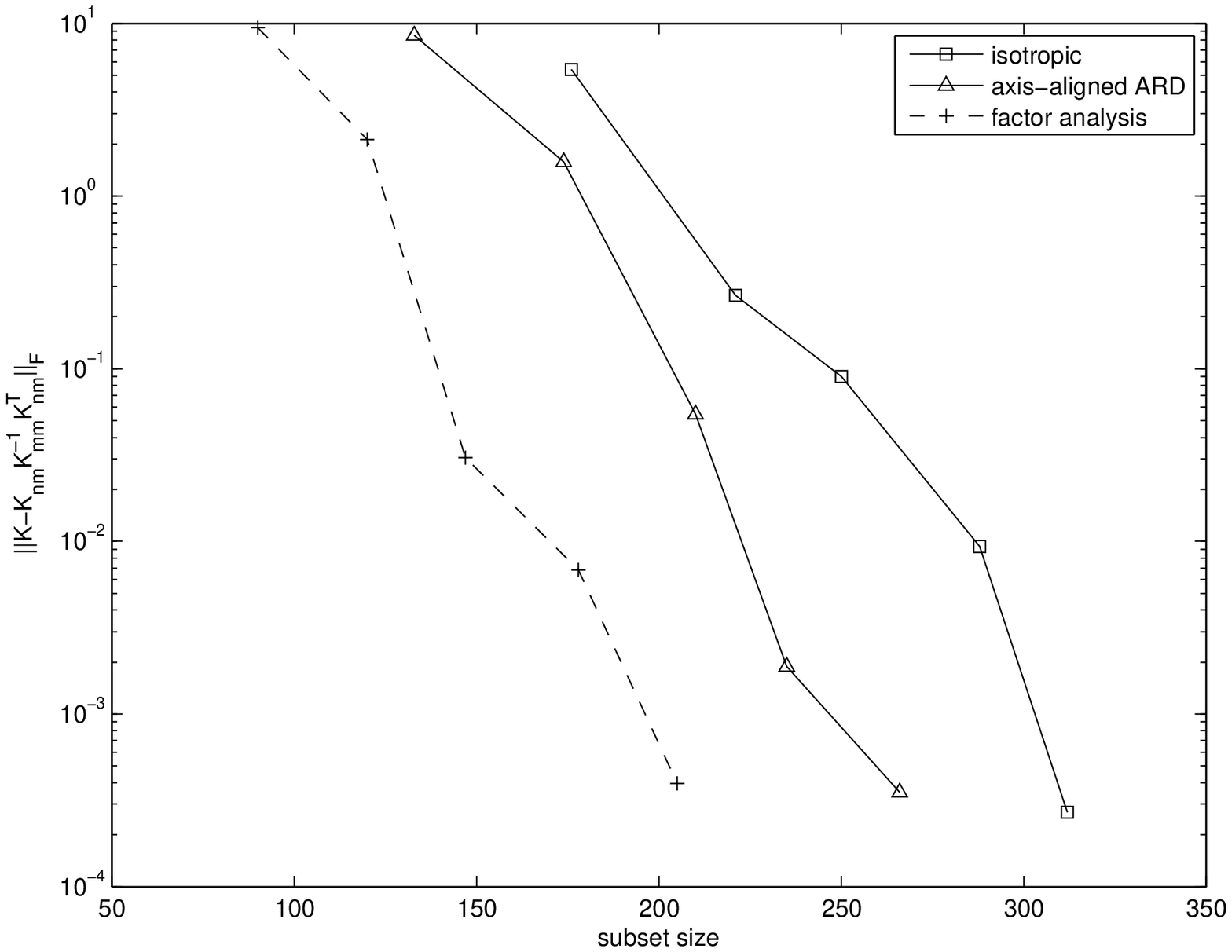}
\end{center}
\caption{Properties of $\bK$ for different choices of $k(\cdot,\cdot)$. 
Left: Eigenspectrum. Center: Number of elements incomplete Cholesky selects for a given 
threshold. Right: Approximation error $\norm{\bK - \tilde\bK}_F$, given the size of the subset.}
\label{fig:3} 
\end{figure}

Additional insight is gained by looking at the eigenspectrum of $\bK$. 
Figure~\ref{fig:3} (left) shows that (I)'s eigenvalues decrease the slowest, whereas
(III)'s decrease the fastest. This has two consequences. First, the
eigenspectrum is intimately related with complexity and generalization
capabilities (see Eq.~\eqref{eq:3.1}) and thus helps explain why (III) 
delivers better prediction performance. Second, the eigenspectrum also 
indicates the effective rank of $\bK$ and strongly impacts our ability
to build an efficient low-rank approximation of $\bK$ using as small a 
subset as possible (see Section~\ref{sec:Computationalconsiderations}).
A small subset in turn is important for computational efficiency because
its size is the dominant factor when we employ the SR-approximation:
both for batch and online learning the operation count depends 
quadratically on the size of the subset (and only linear on the 
number of datapoints). Keeping this size as small as possible without
losing predictive performance is essential. Figure~\ref{fig:3} (center and right)
shows that
in this regard (III) performs best and (I) worst: for example, if we were to 
approximate $\bK$ using SR-approximation with ICD selection at a tolerance 
level of $10^{-1}$, out of our 1000 samples (I) would choose $\sim 175$,
(II) would chose $\sim140$, and (III) would choose $\sim 80$ elements.

\subsection{A 2D gridworld with 1 latent dimension}
\begin{figure}[tb]
\begin{center}
\includegraphics[width=0.32\textwidth]{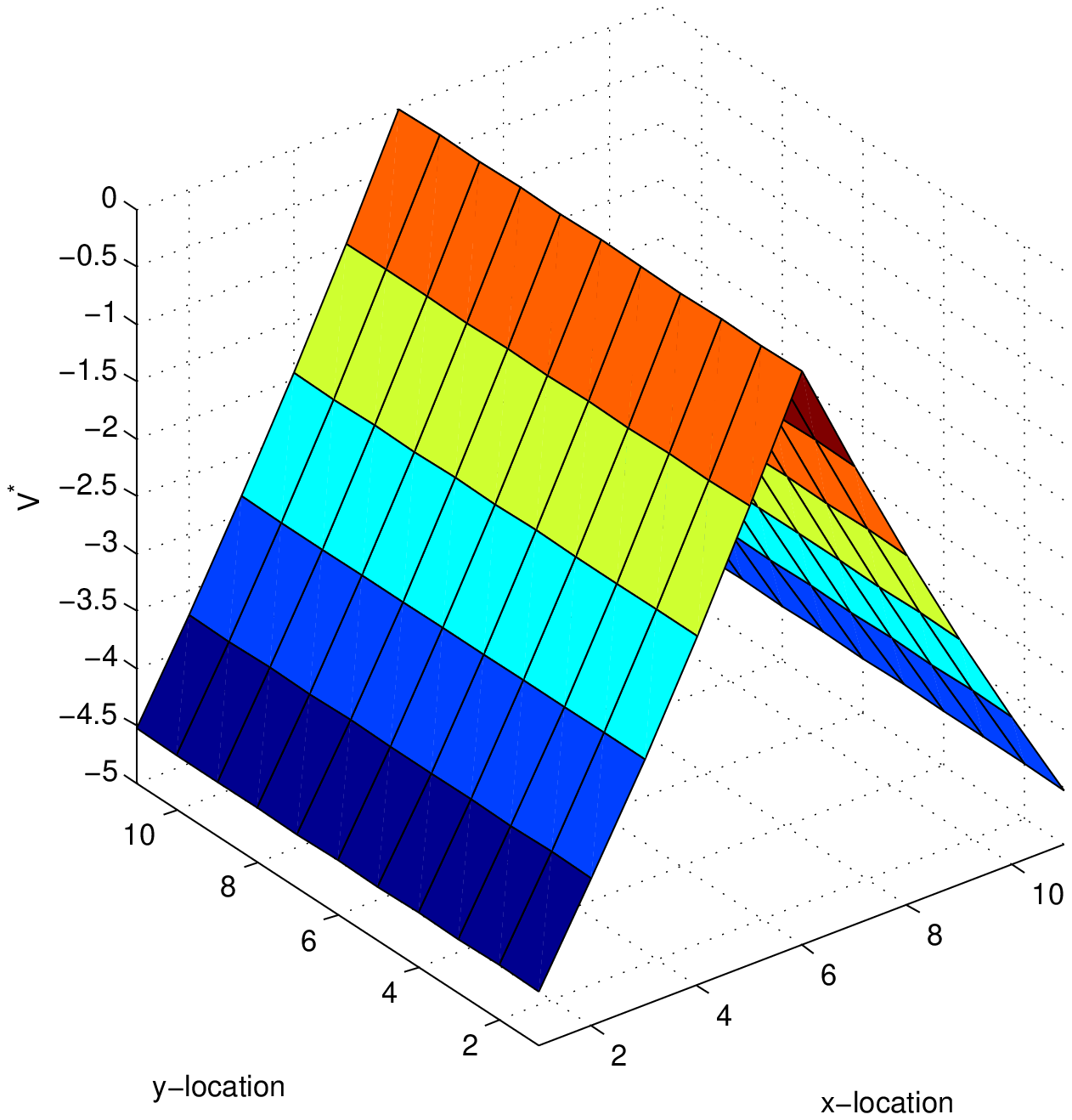}
\includegraphics[width=0.32\textwidth]{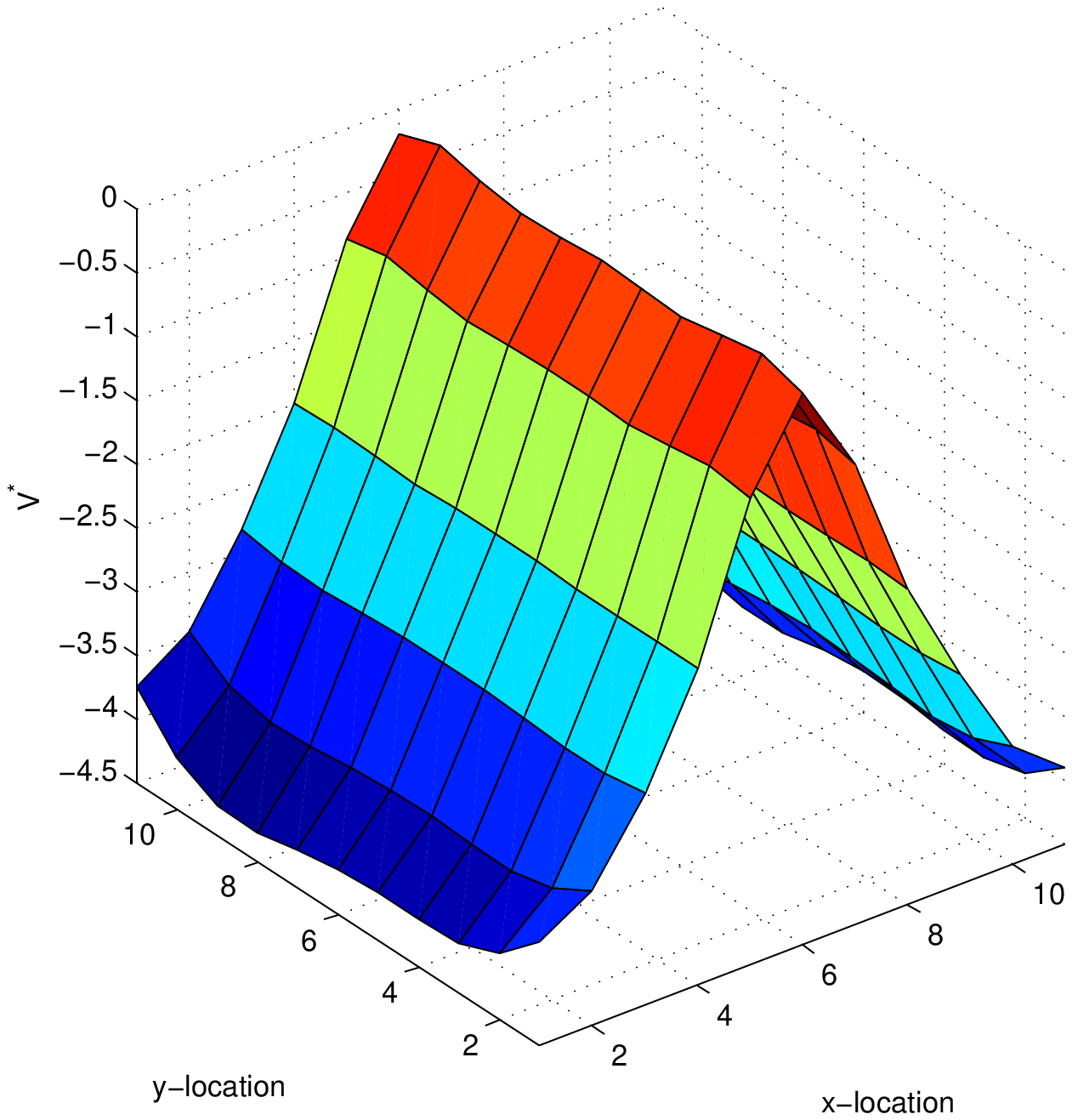}
\includegraphics[width=0.32\textwidth]{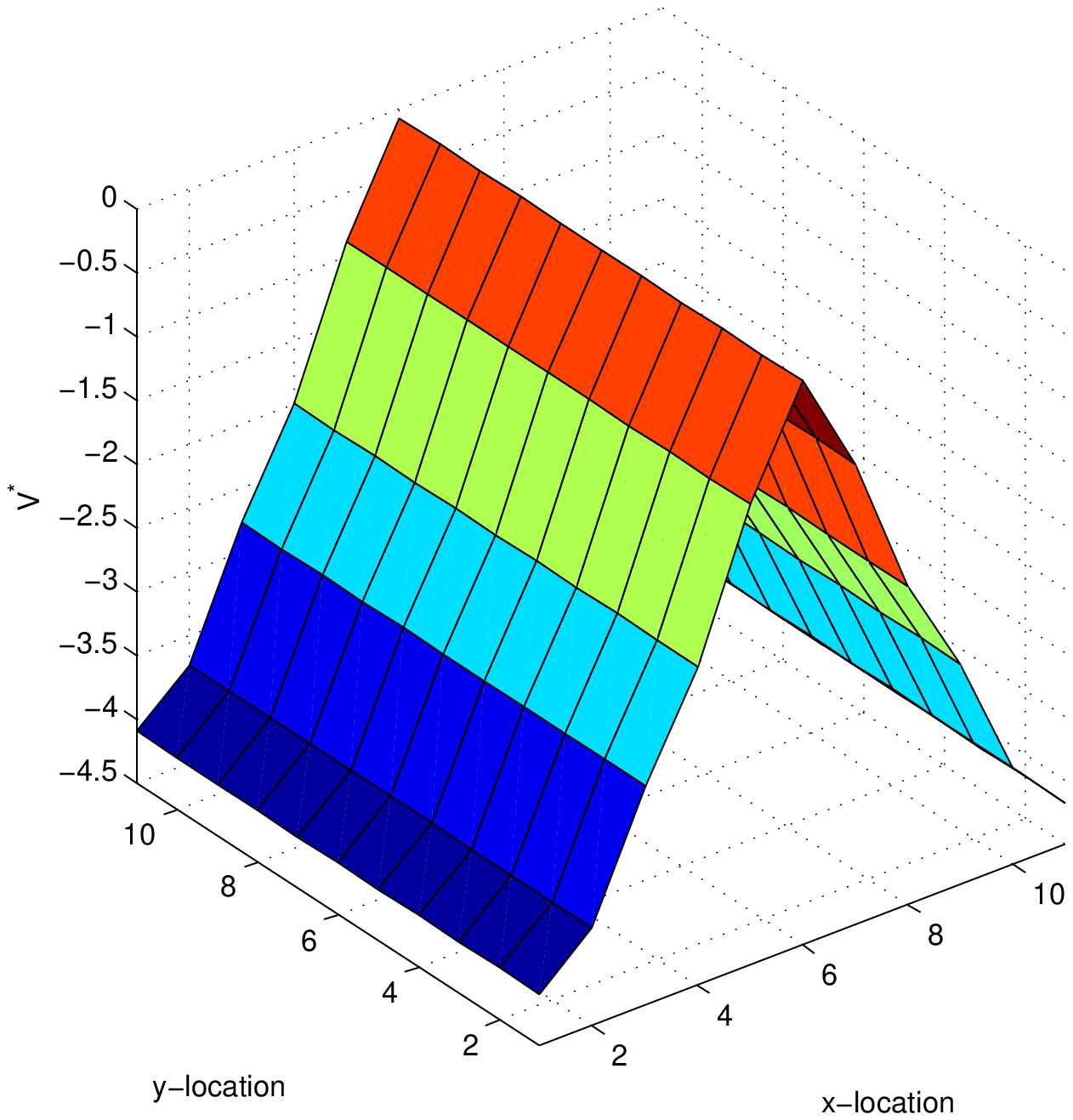}
\end{center}
\caption{Learned value functions for the 2D-gridworld domain. Left: true value function. Note how the y-coordinate is irrelevant for the value. Center: approximation with isotropic covariance. Right: approximation for axis-aligned ARD covariance (after removal of irrelevant input).} 
\label{fig:4}
\end{figure}

\begin{figure}[tb]
\begin{center}
\includegraphics[width=0.49\textwidth,height=4.2cm]{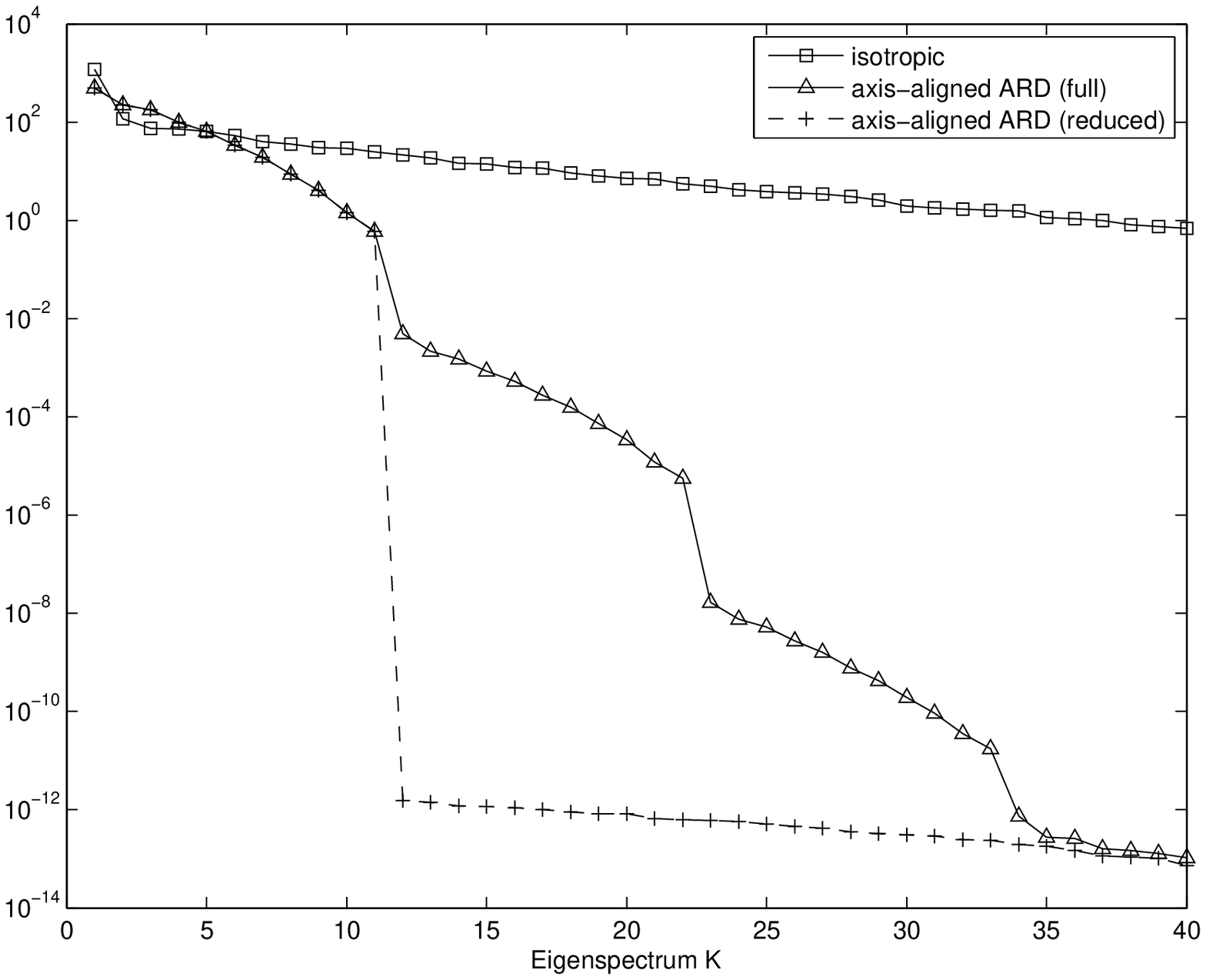}
\includegraphics[width=0.49\textwidth,height=4.2cm]{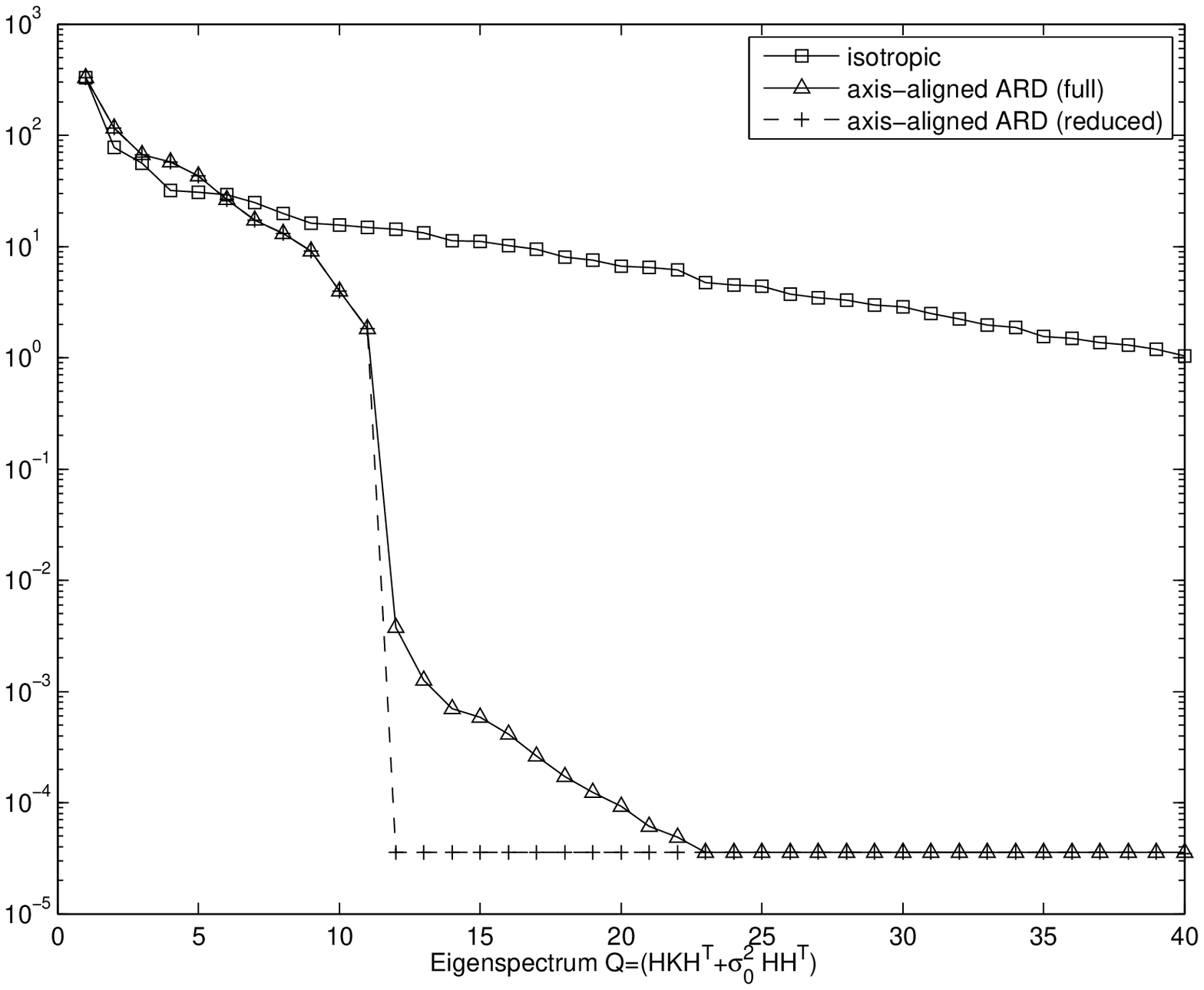}
\end{center}
\caption{Effect of dimensionality reduction on the complexity of the model. Left: Eigenvalues of $\bK$. Right: Eigenvalues of $\bQ$.}
\label{fig:5} 
\end{figure}

To illustrate in more detail how our approach handles irrelevant state variables,
we use a specifically designed 2D gridworld with $11\times11$ states. 
Every step
entails a reward of $-1$ except when being in a state with $x=6$, 
which starts a new episode (teleports to a new random state with
zero reward). We consider the policy that moves left when $x>6$ and right when
$x<6$. In addition, every time we move left or right we will also move randomly
up or down (with 50\% each). The corresponding value function is shown in 
Figure~\ref{fig:4} (left). We generated 500 transitions and applied GPTD with covariance (I) and
(II) with automatic model selection resulting in\footnote{Here we do not include results for (III)
which operates on linear combinations of states and in this scenario would have to find a direction
that is perfectly aligned with the $x$-axis (which is more difficult).} 
\begin{center}
{\small
\begin{tabular}{|l|l|cc||c|}
\hline
                             & \ Hyperparameters $\btheta$ \ & \ Complexity \  & \ Data fit \  & \ $\mathcal L$ (smaller is better)\\
\hline
(I)                          &  $ \ h=2.89$                        & -2378.2 & 54.78 & -2323.4  \\
(II)                         &  $ \ a_1=3.53 \quad a_2=10^{-5} \ $ & -2772.7 & 13.84 & -2758.8  \\
(II) without $y$             &  $ \ a_1=3.53 \quad a_2=0 $         & -2790.7 & 13.84 & -2776.8  \\
\hline
\end{tabular}
}
\end{center}
\noindent
As can be seen from Figure~\ref{fig:4} (center and right), both obtain a very reasonable 
approximation. However,
(II) automatically detects that the $y$-coordinate of the state is irrelevant and 
thus assigns a very small weight to it ($a_2<10^{-5}$). With a uniform lengthscale, (I) is unable
to do that and has to put equal weight on both state variables. As a consequence,
its estimate is less exact and more wiggly (MSE: (I) 0.030, and (II) 0.019).
Additional insight can be gained by looking at the likelihood $\mathcal L$ of the models 
(cf. Eq.~\eqref{eq:3.1}). 
Here we see that (II) has lower complexity (cf. eigenspectrum of $\bQ$ in Figure~\ref{fig:5}),
fits the data better and thus has a higher combined likelihood (note that the values
in the table show the negative log likelihood which we minimize). Moreover, if we completely remove
the $y$ state variable (setting $a_2:=0$), the eigenspectrum of $\bQ$ decreases more rapidly;
thus (II) without $y$ has an even lower complexity while still having the same fit. This indicates
that state component $y$ can be safely ignored in this task/domain. In addition, as was mentioned before, 
the lower effective rank of $\bK$ will also allow us to make more efficient use of SR-based approximations.

\section{Future work}
It should be noted that the proposed framework for automatic feature generation and model selection should primarily be thought of as a practical tool:  despite offering a principled solution to an important problem in RL, ultimately it does not come with any theoretical guarantees (due to some modeling assumptions from GPTD and the way the hyperparameters are obtained). For most practical applications this might be less of an issue, but in general care has to be taken.

The framework can be easily extended to perform policy evaluation over the joint 
state-action space to learn the model-free Q-function (instead of the V-function): 
we just have to choose a different covariance function, taking for example the product 
$k([\bx,a],[\bx',a'])=k(\bx,\bx')k(a,a')$ with $k(a,a')=\delta_{a,a'}$ for problems with 
a small number of discrete actions \cite{mein_keepaway2007}.
This opens the way for model-free policy
improvement and thus optimal control via approximate policy iteration. Our next step then is  
to apply this approach to real-world high-dimensional control tasks, both in batch settings
and hybrid batch/online settings; in the latter case exploiting the gain in
computational efficiency obtained through model selection to improve 
\cite{engel2005rlgptd}. 
\section*{Acknowledgments}
This work has taken place in the Learning Agents Research
Group (LARG) at the Artificial Intelligence Laboratory, The University
of Texas at Austin.  LARG research is supported in part by grants from
the NSF (CNS-0615104), DARPA
(FA8750-05-2-0283 and FA8650-08-C-7812), the Federal Highway
Administration (DTFH61-07-H-00030), and General Motors.

\bibliographystyle{plain}

\end{document}